\pdfoutput=1

\documentclass[11pt]{article}

\usepackage[]{ACL2023}

\usepackage{times}
\usepackage{latexsym}

\usepackage[T1]{fontenc}

\usepackage[utf8]{inputenc}

\usepackage{microtype}

\usepackage{inconsolata}

\usepackage{graphicx}

\usepackage{enumitem}

\usepackage[subtle]{savetrees}
\usepackage{subcaption}
\usepackage{float}
\usepackage{gb4e}

\title{Investigating Syntactic Biases in Multilingual Transformers\\ with RC Attachment Ambiguities in Italian and English }

 \author{Michael Kamerath \and Aniello De Santo\\
         University of Utah}

\begin{document}
\maketitle

\begin{abstract}
This paper leverages past sentence processing studies to investigate whether monolingual and multilingual LLMs show human-like preferences when presented with examples of relative clause attachment ambiguities in Italian and English.\
Furthermore, we test whether these preferences can be modulated by lexical factors (the type of verb/noun in the matrix clause) which have been shown to be tied to subtle constraints on syntactic and semantic relations.\ 
Our results overall showcase how LLM behavior varies interestingly across models, but also general shortcomings of these models in correctly capturing human-like preferences.\
In light of these results, we argue that RC attachment is the ideal benchmark for  cross-linguistic investigations of LLMs' linguistic knowledge and biases.
\end{abstract}

\section{Introduction}

The ubiquitousness of Large Language Models (LLMs), as they get incorporated in more day-to-day applications, makes it crucial to investigate the ways in which their behavior on specific linguistic input resembles or differs from that of humans --- an approach which can contribute to understanding the type of linguistic knowledge they capture.\@

In this sense, a recent but already classical line of work has focused on evaluating neural models' predictions on fine-grained syntactic phenomena/constructions, in order to probe whether the models have learned knowledge about the specific structural characteristics of (a) language \citep{linzen2016assessing,marvin2018targeted,gauthier2020syntaxgym,warstadt2019neural,warstadt2020learning,warstadt2020blimp,sartran2022transformer,newman2021refining,jumelet2024black,arora2024causalgym}.
In fact, this type of comparison might allow us to leverage psycholinguistic theories to gain insight into the opaque (learned or architectural) biases of LLM \citep{linzen2021syntactic,futrell2019neural,ettinger2020bert}.\@

While a majority of past work has focused on evaluating LLMs' syntactic knowledge in terms of their ability to distinguish grammatical and ungrammatical constructions, an important component of human sentence comprehension is ambiguity resolution \citep{altmann1998ambiguity,gibson1998constraints}.\@
In particular, it is worth investigating how neural models handle multiple \emph{simultaneously correct interpretations} for a single sentence in the absence of disambiguating cues/context \citep{davis2020recurrent,bhattacharya2022sentence,liu2023we,zhou2024tree}.\

Consider the case of a relative clause (RC) (\textit{that was running}) following a complex noun phrase (\textit{son of the doctor}), as in (\ref{ex:rcAmbiguity}):
\begin{exe} 
    \ex I saw the \underline{son} of the \underline{doctor} that was running.
    \label{ex:rcAmbiguity}
\end{exe}
There are two possible interpretations of this sentence: the interpretation in which the RC modifies \textit{the doctor} is usually referred to as low attachment (LA), while the case of the RC modifying \textit{the son} is referred to as high attachment (HA). 

Famously, human preferences for HA or LA vary both individually and cross-linguistically, and are affected by a variety of syntactic and semantic factors  \cite{cuetos1988cross,de1993some,mitchell1990reading,miyamoto1998relative,maia2007early,abdelghany1999low}.\
Moreover, some of these factors (e.g., the type of verb used in the matrix clause of the sentence) seem to be tied to subtle syntactic differences in each language \cite[a.o.]{cinque1992pseudo,grillo2015highs,GRILLO2014156}.

RC attachment ambiguities thus present an interesting way of probing LLMs' syntactic knowledge and behavior.\@
In fact, investigating LLMs' performance over ambiguous sentences cross-linguistically might provide crucial insights into the kind of linguistic biases available to these models through their training data, and the properties of the models tied to architectural choices \citep{davis2020recurrent,li2024incremental}.\
As differences in the frequency of HA vs. LA structures have been argued to account for the cross-linguistic variation of RC preferences at least to some degree, it seems reasonable that LLM models would be able to replicate (some of) these patterns.\
However, RC attachment seems to be understudied in the LLM syntactic evaluation literature \citep{davis2020recurrent,ISSA2024425}.\

Here, we aim to add to this scarce literature, and evaluate a variety of LLMs to determine their disambiguation strategies for RCs in Italian and English.\
We compare Italian to English since the two languages have some shared structural properties (e.g., SVO, post-nominal RCs), but differ in RC interpretation: modulo other variables, English speakers generally exhibit a LA RC preference while Italian speakers a HA one \citep{frazier1983processing,cuetos1988cross,de1993some}.\
Additionally, Italian and English speakers respond differently to other variables affecting RC attachment, which have been argued to be also captured by some multilingual LLMs \citep[e.g., type of matrix verb;][]{grillo2015highs,GRILLO2014156,henot2023language}.\
Therefore, building on the psycholinguistics and LLM literature on RC attachment, here we ask:
\begin{enumerate}
    \item whether monolingual and multilingual LLMs tested on Italian and English show any type of attachment preference when presented with ambiguous RCs;
    \item whether these preferences conform to those of Italian/English speakers;
\item whether these preferences show sensitivity to fine-grained structural information modulated by properties of the matrix clause.
\end{enumerate}

\section{Related Work}
\label{sec:Related}

The cross-linguistic variability of attachment preferences for ambiguous RCs has been a focus of many psycholinguistics debates, due to its direct relevance to questions about the mechanisms guiding human sentence processing \citep[a.o.]{frazier1983processing,cuetos1988cross,de1993some,gibson1998constraints,grillo2015highs,hemforth2015relative, Lee2024ItalianRC}.\

Famously, when presented with a globally ambiguous sentence like in (\ref{ex:rcAmbiguity}), and in the absence of a disambiguating context, English speakers tend to prefer a LA interpretation: an interpretation in which the RC gives us information about (\emph{modifies)} the second noun of the preceding complex noun phrase \citep{frazier1983processing,cuetos1988cross}.\
This LA preference is well attested in other languages, for example in Mandarin Chinese and Arabic \cite{shen2006late,abdelghany1999low,ehrlich1999low}.\
In turn, a preference for the RC modifying the first noun --- a HA interpretation --- has been found in languages like Italian, Spanish, or Dutch \citep{cuetos1988cross,de1993some,brysbaert1996modifier,frenck2000romancing,mitchell2000modifier}.\
Beyond these broader preferences at the language level, multiple factors have been shown to affect RC preferences across languages --- for instance, referentiality of the modified nouns, lexical and structural frequency, semantic or pragmatic plausibility, length and structural position of the RC, implicit prosody, individual working memory differences, or task type \citep{de1993some,macdonald1994lexical,gilboy1995argument,ferreira2003misinterpretation,fernandez2003bilingual,swets2008underspecification,acuna2009animacy}.\

Recently, it has been argued that one important predictor of attachment disambiguation in Italian RCs is whether the verb in the main clause is non-perceptual (\textit{marry, know, cook, etc}) or perceptual (\textit{observe, hear, smell, etc}).
When other semantic and syntactic aspects are controlled for, RCs of sentences containing non-perceptual verbs lead to a LA preference while perceptual verbs lead to a HA preference \cite{GRILLO2014156, Lee2024ItalianRC}.\
More generally, reviewing past literature on RC attachment preferences in so-called HA languages, \citet{GRILLO2014156} have related this verb-type sensitivity to the availability to a subtle structural ambiguity at the complementiser, beyond the classic LA RC vs.\ HA RC choice.\
Some languages allow for a construction known as a Pseudo-Relative Clause (PRs), which is string-identical to RCs but different at the semantic, syntactic, and prosodic levels \citep[a.o.]{cinque1992pseudo,GRILLO2014156,aguilarSpanishNoDifferent2021}.\
In particular, instead of providing information about the entity (noun) that is modified, PRs denote direct perception of events and are thus only compatible with some specific subclasses of verbs (e.g., \textit{photograph, record}) in the matrix clause (perception verbs, introducing events).\
Importantly, PRs are only compatible with what looks like a HA interpretation, leading to an apparent HA preference with verbs that license them.\
This hypothesis has found general experimental support in a variety of languages including Italian \cite{GRILLO2014156, Lee2024ItalianRC} and Spanish \cite{aguilarSpanishNoDifferent2021}.\

RC attachment thus seems to offer ways to explore the sensitivity of LLMs to a variety of important structural and semantic features within and, crucially, across languages.\
As mentioned, starting with \citet{linzen2016assessing}, there has been a fruitful line of research using psycholinguistic tasks to explore neural models' knowledge of different lexical, structural, and semantic linguistic properties \citep{marvin2018targeted, gauthier2020syntaxgym,gauthier-etal-2022-preuve, warstadt2019neural, warstadt2020learning, warstadt2020blimp, sartran2022transformer,newman-etal-2021-refining,jumelet-etal-2021-language,arora-2022-universal,gulordava2018colorless,sinclair2021syntactic,goldberg2019assessing,wilson2023subject} --- and to evaluate whether model behavior resembles the performance of humans tested on similar tasks/constructions \citep{sinha2021unnaturallanguageinference,futrell2019neural,ettinger2020bert}.\

While some work probing LLMs' ability to deal with (different types of) ambiguity exists \citep{van2018modeling,bhattacharya2022sentence,liu2023we,zhou2024tree,li2024incremental}, little attention has been paid to the phenomenon of RC attachment in absence of a disambiguating context.\
In this sense, \citet{davis2020recurrent} analyzed the ability of LSTMs to learn RC attachment preferences in English and Spanish.\@
They showed that LSTMs preferred English-like attachment (LA) in both English and Spanish.\ 
More recently, \citet{ISSA2024425} tested RC attachment in Arabic with a variety of transformer models \citep{vaswani2017attention}, using a zero-shot prompting method.\
They showed significant variability across model architectures, with some models' behavior being in line with the attachment preferences reported for Arabic speakers, while others showing no preference at all.\
Furthermore, going back to our discussion of linguistic factors that modulate RC preferences, \citet{henot2023language} has shown that monolingual and multilingual transformer architectures exhibit some sensitivity to PR-availability in French.\
However, this work evaluated PR-related properties only in contexts outside of ambiguous RC, and it thus unclear whether they would modulate an LLM's choice of attachment.\

In sum, the complex interaction between RC attachment and other syntactic/semantic factors opens an exciting set of possibilities for the cross-linguistic evaluation of LLMs' behavior.\
In what follows, building on the results of \citet{davis2020recurrent} and \citet{henot2023language}, we focus on evaluating a set of monolingual and multilingual models on the patterns of RC-attachment and PR-sensitivity reported in the psycholinguistic literature for Italian and English \citep{GRILLO2014156,grillo2015highs,Lee2024ItalianRC}.\

\section{Italian Experiment}
\label{sec:Experiment1}

\begin{table}
\centering
\resizebox{1\linewidth}{!}{  
\begin{tabular}{|c|c|c|}
    \hline
    Model Name & Language & Reference \\\hline
    GePpeTto & Italian & \citet{de2020geppetto} \\\hline
    Alberto & Italian & \citet{polignano2019alberto} \\\hline
    bert-base-multilingual-cased & multilingual & \citet{devlin2019bertpretrainingdeepbidirectional} \\\hline
    xlm-mlm-17-1280 & multilingual & \citet{conneau2019cross} \\\hline
    xlm-roberta-large & multilingual & \citet{conneau2020unsupervisedcrosslingualrepresentationlearning} \\\hline
\end{tabular}
}
\caption{Italian and Multilingual Models in this paper.}
\label{tab:experimentLLMs}
\end{table}

As mentioned, past literature suggests that in languages that allow for PRs (e.g., Italian) --- when controlling for other linguistic factors ---  if the matrix verb is perceptual a PR interpretation takes precedence, resulting in a HA preference.\
Otherwise, a LA preference is observed.\

\citet{GRILLO2014156} tested this prediction by evaluating Italian participants' behavior when exposed to globally ambiguous sentences containing a complex noun phrase followed by an RC.\
Sentences varied over the type of verb used in the matrix clause (perceptual/stative).\
As predicted, participants showed an HA preference only with perceptual verbs, and exhibited an ``English-like'' LA preference with stative verbs.\

Here, we exploit this design to explore whether the type of matrix verb in Italian sentences affects LLM attachment preferences.\
In the past, a common evaluation technique has been to check whether a model assigns a higher probability to a grammatical sentence compared to an ungrammatical one \citep{linzen2016assessing,gulordava2018colorless}.
However, here we are interested in probing an LLM's preference in choosing one grammatical interpretation over another equivalently grammatical one, in the absence of other disambiguating factors (e.g.,\ context).\
To do so, instead of using the globally ambiguous sentences of \citet{GRILLO2014156}, we follow \citet{davis2020recurrent} and adopt sentences that are temporarily ambiguous.\
Specifically, we adopt a modification of the \citet{GRILLO2014156}'s stimuli presented by  \citet{Lee2024ItalianRC}.\

This work follows a $2\times2$ design, in which quartets of sentences vary across two dimensions: Verb Type and Attachment Type.\
As in \citet{GRILLO2014156}, sentences include a main verb which is either perceptual (\emph{heard}) or stative (\emph{worked with}) and a complex noun phrase (\emph{the grandma of the girls}) followed by an RC.\
Items are disambiguated towards HA or LA based on singular/plural agreement between one of the nouns in the matrix clause (\emph{grandma/girls}), and the embedded verb.\@
This is possible since Italian differentiates singular/plural morphology explicitly on the main verb (see the examples in \ref{exe:ItalianStimuli}).\

\begin{table}[htbp]
    \centering
    \begin{tabular}{|c|c|c|}
    \hline
        Sentence & Verb Type & Attachment \\\hline
        a & perceptual (P) & HA \\\hline
        b & perceptual (P) & LA \\\hline
        c & non-perceptual (N) & HA \\\hline
        d & non-perceptual (N) & LA \\\hline
    \end{tabular}
    \caption{Summary of $2\times2$ design in the Italian Experiment.}
    \label{tab:ItalianStimuli}
\end{table}

\begin{exe}
    \ex Italian Stimuli \citep{Lee2024ItalianRC}
		\begin{xlist}
            \ex \gll Maria sentí la nonna delle ragazze che gridava gli insulti \\
          Maria heard-\sc{3SG} the grandma {of the} girls who screaming-{\sc 3SG} the insults\\
            \glt ``Maria heard the grandma of the girls who was screaming the insults''

            \ex \gll Maria sentí la nonna delle ragazze che gridavano gli insulti \\
            Maria heard-\sc{3SG} the grandma {of the} girls who screaming-{\sc 3PL} the insults\\
            \glt ``Maria heard the grandma of the girls who were screaming the insults''

            \ex \gll Maria lavoró con la nonna delle ragazze che gridava gli insulti \\
            Maria worked-\sc{3SG} with the grandma {of the} girls who screaming-{\sc 3SG} the insults\\
\glt ``Maria worked with the grandma of the girls screaming who were screaming the insults''    
            \ex \gll Maria lavoró con la nonna delle ragazze che gridavano gli insulti \\
                        Maria worked-\sc{3SG} with the grandma {of the} girls who screaming-{\sc 3PL} the insults\\
            \glt ``Maria worked with the grandma of the girls who were screaming the insults''
		\end{xlist}
        \label{exe:ItalianStimuli}
\end{exe}

We use \citet{Lee2024ItalianRC}'s items, which include 24 sentence sets for a total of ninety-six sentences.\@
Each set contains $4$ sentences varying across the two dimensions mentioned above, as summarized in Table \ref{tab:ItalianStimuli}.\
In line with the models tested for French by \citet{henot2023language}, we test two Italian-only models, and three multilingual models (GePpeTto; AlBerto; bert-base-multilingual-cased; xlm-mlm-17-1280; xlm-roberta-large; see Table \ref{tab:experimentLLMs}).\

Following \citet{davis2020recurrent}, we evaluate LLMs using information-theoretic surprisal \citep{hale2001probabilistic, levy2008expectation}, which is usually defined as the inverse log probability assigned to a word in a sentence given its preceding context.\
Fixed verb-type, our stimuli include sentence pairs that are string identical until the singular/plural features on disambiguating verb.\@
Thus, for each item in the dataset, we compute surprisal at the embedded verb using the \texttt{minicons} library \cite{misra2022minicons}.\
In terms of qualitative interpretation, when comparing sentence types a low surprisal value for a LA item compared to its paired HA item would indicate a LA preference, and viceversa.\
Verb-type sensitivity would show these high/low surprisal values at the embedded verb also modulated by the properties of the matrix verb.\

For each LLM, we fit a linear mixed-effect model using Surprisal at the embedded verb as the dependent variable, and Verb Type and Attachment Type as fixed effects.\
We also include a random slope for set, in order to account for lexical variation across sentence quartets.\footnote{\texttt{Surprisal $\sim$ Verb Type + Attachment Type + Verb Type*Attachment Type + (1|set)}}\
All analyses were performed using R Statistical Software \citep[R version 4.4.1;][]{Rpackage}, 2024), using the lme4 package \citep[version 1.1.35.5;][]{bates2015package}.

\begin{figure*}[htbp]
    \centering
    \includegraphics[width=1\linewidth]{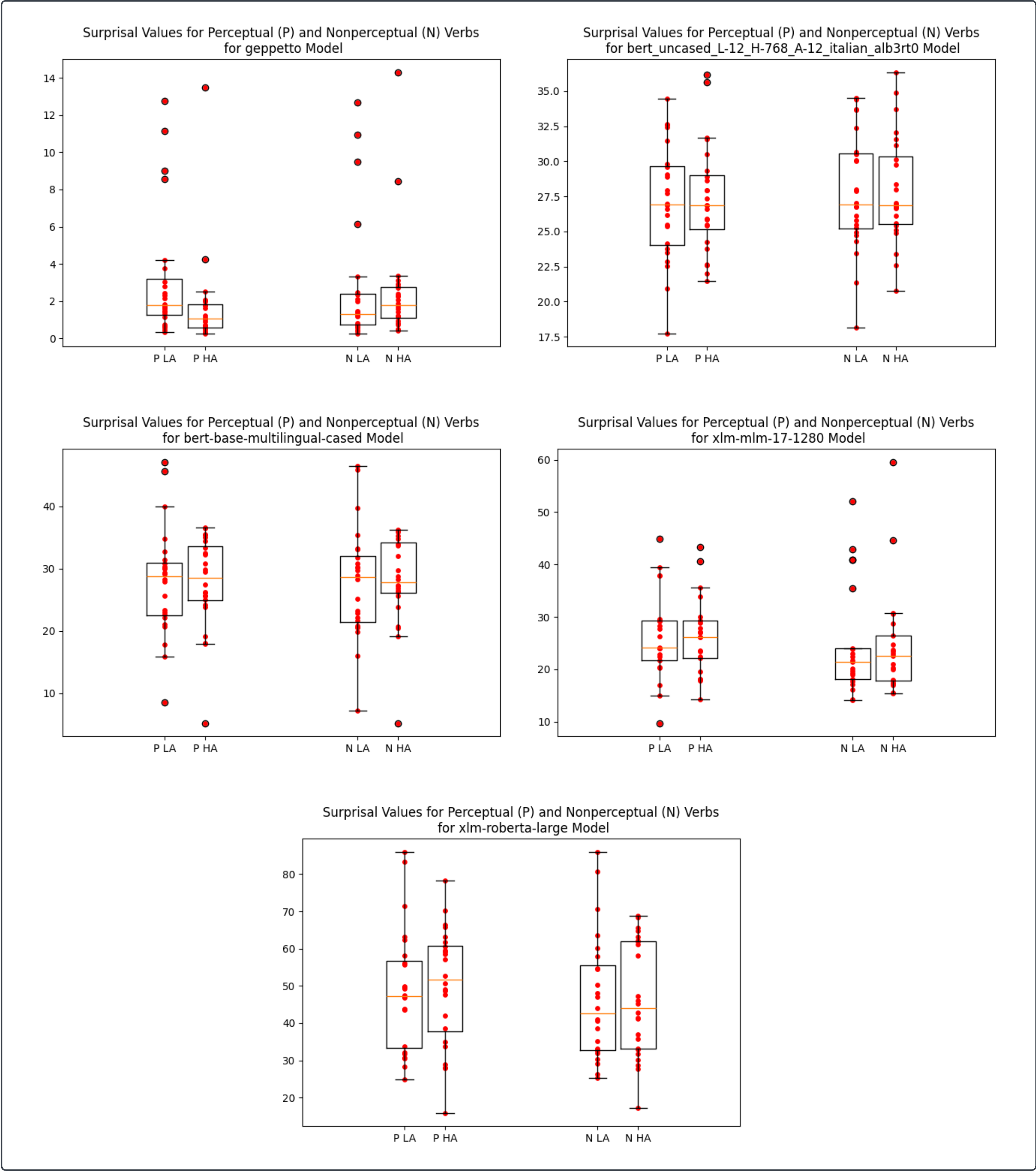}
    \caption{Surprisal values by condition, for each one of the models tested in the Italian Experiment.}
    \label{fig:SurprisalScores}
\end{figure*}

While some trends arise from qualitative pairwise comparisons (cf. Appendix \ref{sec:appb}), statistical analyses show no significant attachment or verb type effects, nor their interaction, for any of the LLM tested.\@
These results can be interpreted as the absence of an attachment preference (in line with Italian speakers or not), and a lack of sensitivity to verb-type properties, in both the Italian-only and the multilingual models (see Figure \ref{fig:SurprisalScores}).\footnote{LMER output for each of the statistical models fit in this paper can be find in Appendix \ref{sec:appa}.}

\section{English Experiments}

Beyond a PR-based account of attachment preferences in Italian, \citet{grillo2015highs} observe that a pragmatic explanation could also be viable, since PR availability co-varies in Italian with semantic properties of the matrix verb (i.e., implicit causality).\
To test this hypothesis, they conducted an English study probing similar variables modulating RC attachment as those manipulated in the Italian studies discussed above.\

While not a PR-language, English allows for structures that are interpretatively similar to PRs --- eventive small clauses (SC) --- and are licensed by the same verb-types as PRs in Italian.\@
However, since in English SCs are not string equivalent to RCs including an explicit complementiser, PR-related verb-type effects should not arise with RC sentences independently of the type of matrix verb used.\
In an offline questionnaire, \citet{grillo2015highs} then show that English participants consistently prefer a LA interpretation, even though they observe a small HA boost in the SC-licensing/Perceptual verb condition.\ 
They argue that these results are incompatible with a pragmatic account of the Italian findings.\

This experiment offers us a way to further probe factors affecting RC attachment strategies in LLMs with a direct cross-linguistic comparison of the manipulated variables.\
Additionally, as implicit causality has been explored in LLM literature to somewhat conflicting results, this stimulus set up might lead to broader insights into LLMs' sensitivity to semantic/pragmatic variables \citep{kankowski2025implicit,kementchedjhieva2021john}.\

We thus aim to adopt the stimuli and design of \citep{grillo2015highs} for the LLMs tested here.\
In addition to the verb-type manipulation of the Italian experiments, \citet{grillo2015highs} also modulate the type of nominal used as the first noun in the complex noun --- either licensing a SC or not (\emph{heard} vs. \emph{scream}).\@
This noun-type manipulation implies testing RCs following a complex noun-phrase in the subject position of the main sentence, compared to the object modifying RCs used when manipulating verb type (Example \ref{exe:NomVerb}).

\begin{exe}
    \ex 
		\begin{xlist}
            \ex Kelly heard the grandma of the girl that was screaming.
            \ex The sounds of the grandma of the girl that was screaming is annoying.
		\end{xlist}
        \label{exe:NomVerb}
\end{exe}

Therefore, these English stimuli allow us to investigate an additional structural factor potentially affecting LLMs.
Since we will depart from the psycholinguistic study in again using disambiguated RCs over globally ambiguous ones, we split \citep{grillo2015highs}'s experiment into two: Experiment 1 will test the effect of verb-type in English, while Experiment 2 will text the effects of noun-type/RC position.\
For consistency with the Italian experiment, we test on these English stimuli the three multilingual models evaluated in the previous section.\

\begin{figure*}[htbp]
    \begin{subfigure}[t]{0.5\textwidth}
\centering
    \includegraphics[width=1.0\linewidth]{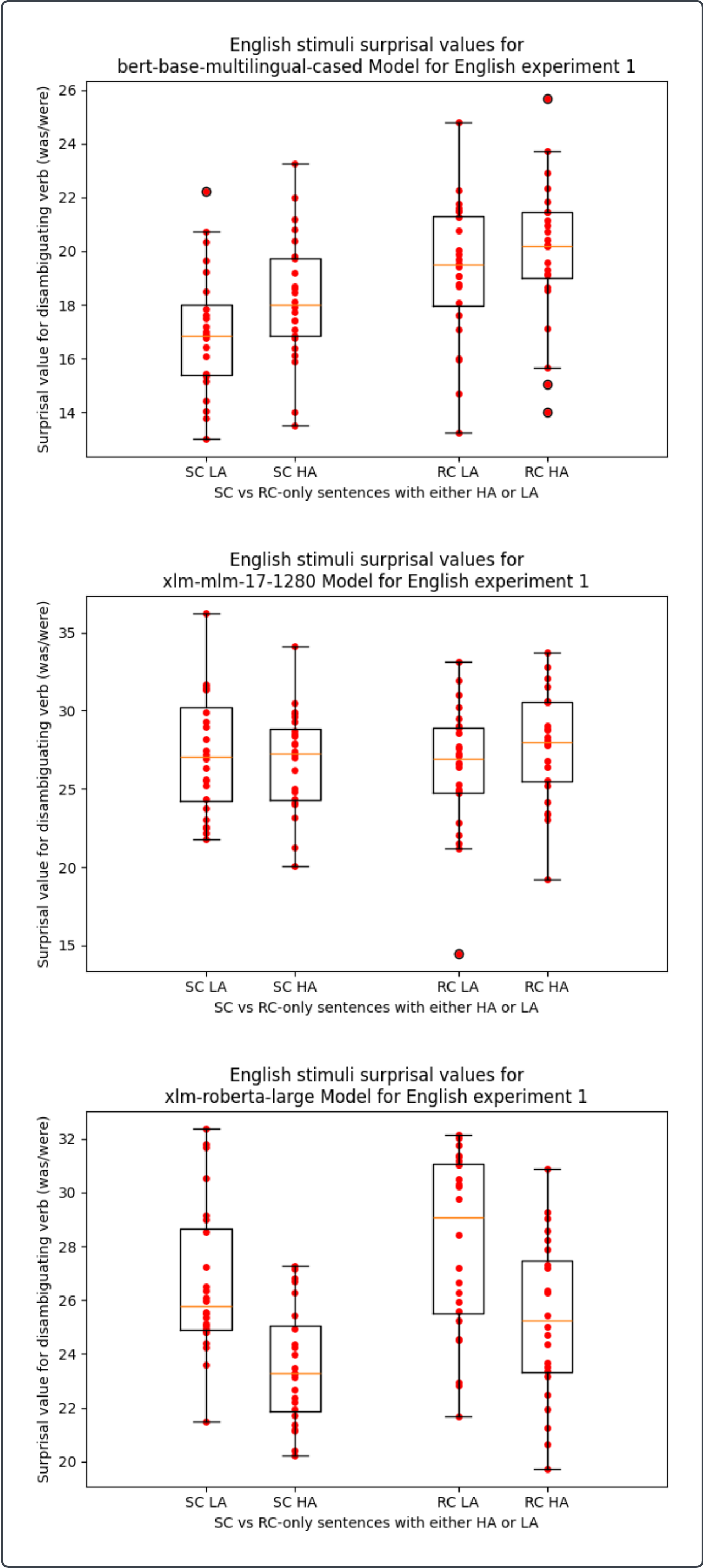}
    \caption{}
    \label{fig:E1}
       \end{subfigure}%
    ~ 
        \begin{subfigure}[t]{0.5\textwidth}
\centering
    \includegraphics[width=1.0\linewidth]{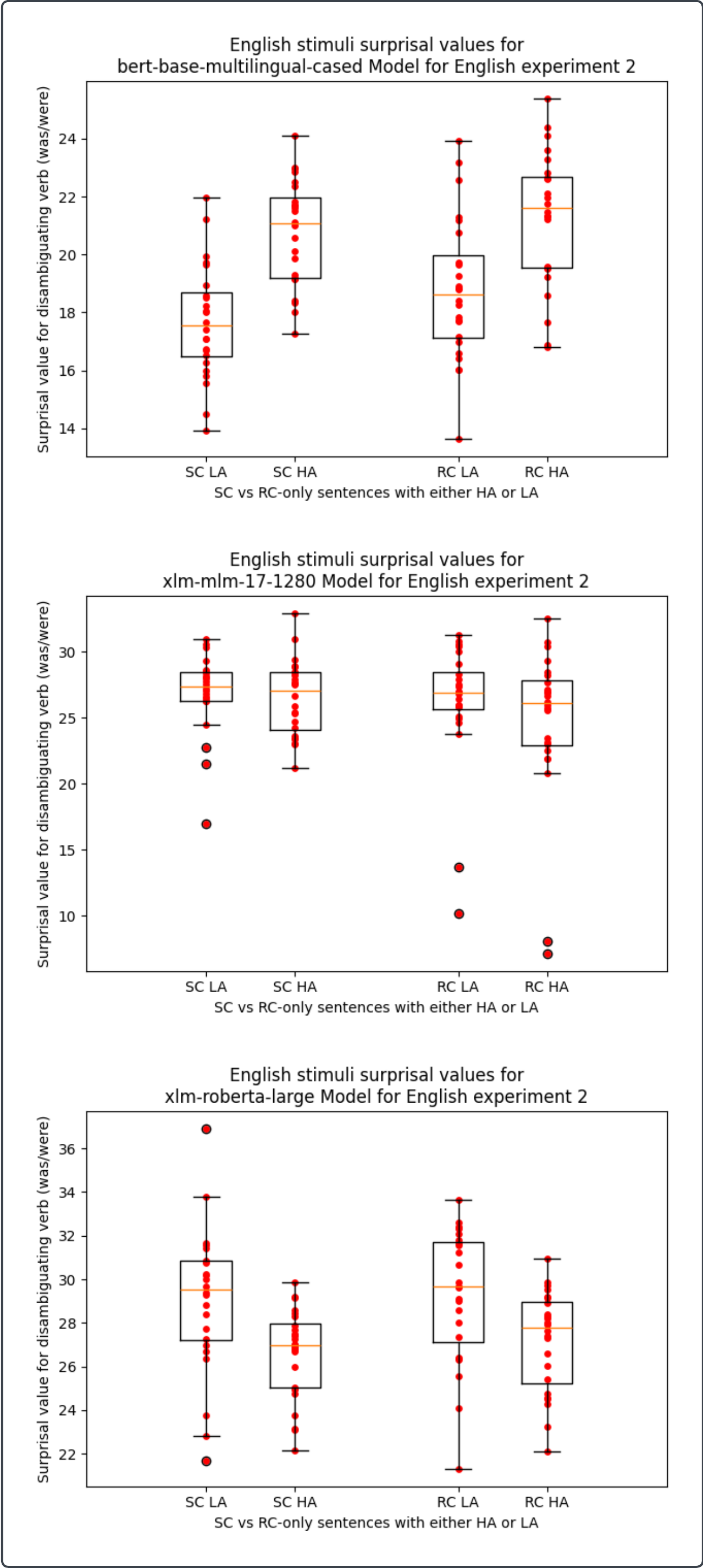}
       \caption{}
    \label{fig:E2}
      \end{subfigure}%
            \label{fig:EnglishFull}
                \caption{Surprisal values by condition, for each one of the models tested in the English Experiment 1 (a) and Experiment 2 (b).}
  \end{figure*}

\subsection{Experiment 1: Verb-Type Effects}
\label{subsect:EnglishExperiment1}

First, we investigate Verb Type effects in English, using stimuli adapted from the first experiment in \cite{grillo2015highs}.\@
These include sets of four lexically matched items holding all properties of a sentence constant except for the matrix verb, which is either a RC-only verb or a SC-licensing verb (see \ref{exe:EnglishVerbStimuli}).\
\citet{grillo2015highs} report that human participants tested on these stimuli showcase a general preference for LA, but a slight HA boost in the SC-licensing condition.

\begin{table}[htbp]
    \centering
    \begin{tabular}{|c|c|c|}
    \hline
        Sentence & Verb Type & Attachment \\\hline
        a & RC-only & HA \\\hline
        b &  RC-only & LA \\\hline
        c & SC & HA \\\hline
        d & SC & LA \\\hline
    \end{tabular}
    \caption{Summary of $2\times2$ design in the English Experiment 1.}
    \label{tab:ENG1Stimuli}
\end{table}

\begin{exe}
    \ex English Exp. 1 Stimuli \citep{grillo2015highs}
		\begin{xlist}
            \ex Jim saw the son of the doctors that was having dinner.
            \ex Jim saw the son of the doctors that were having dinner.
            \ex Jim shares the house with the son of the doctors that was having dinner.
            \ex Jim shares the house with the son of the doctors that were having dinner.
		\end{xlist}
        \label{exe:EnglishVerbStimuli}
\end{exe}

Similarly to the Italian experiment, in our evaluation all items are modified  to disambiguate LA/HA based on singular/plural agreement on the embedded verb.\@
Because of the properties of English, this disambiguation happens over an auxiliary verb (\emph{was/were}) instead of directly on the embedded verb --- see Table \ref{tab:ENG1Stimuli} for a summary of the main properties of the experimental items.\@
The experimental stimuli included twenty-four sets, for a total of ninety-six sentences. 

Again, we fit a linear mixed-effect model using Surprisal at the embedded verb as the dependent variable, and Verb Type and Attachment Type as fixed effects.\
Compared to the Italian Experiments, results here are more mixed (see Figure \ref{fig:E1} and Appendix \ref{sec:appa}).\

For the bert-base model, we found a significant Verb Type effect, consistent with surprisal values being generally lower in the SC-licensing verb condition that in the RC-only condition.\@
These differences are independent of Attachment Type, although with SC verbs we observe a (non-significant) trend in favor of the LA condition ---  which is line with the known LA preference in English, but somewhat in contrast with what \citet{grillo2015highs} found with human participants.\
No significant effects were found with the xlm model, but there were marginal effects of Attachment Type and of the Verb Type/Attachment Type interaction.\
The xlm model's results do trend towards lower surprisal for LA in the RC-only condition (Figure \ref{fig:E1}).\ 
While this trend does not result in a statistically significant difference, among all models tested this pattern is qualitatively the most in line with the data from human participants (see also Appendix \ref{sec:appb}).\
Finally, for the roberta model we found significant Very Type and Attachment Type effects, but no interaction effects.\
Again, surprisal values in the SC condition are lower independently of Attachment Type (Figure \ref{fig:E1}).\
Additionally, surprisal values for HA items are significantly lower than those of  LA items (thus indicating a HA preference).\
In fact, qualitatively it seems that the roberta model prefers HA items in almost every set --- again in contrast with the pattern of preferences usually reported for human English participants (see  Appendix \ref{sec:appb}).\

\subsection{English Experiment 2: Noun-Type Effects}
\label{subsect:EnglishExperiment2}

In a second experiment, we leverage the stimuli in the nominal condition of \citet{grillo2015highs}'s first experiment.\@
This condition compares nominals that license SC (i.e., compatible with the description of an event) to  nominals that are only compatible with RCs.\@
As mentioned above, the nominal condition is also designed so that the complex noun phrase (and thus the following RC) occupies the subject position of the matrix clause (as in \ref{exe:EnglishNounStimuli}).\@
On these stimuli, \citet{grillo2015highs}' English participants show a LA preference, but no noun-type effect.\@
For our LLM tests, we again modify all times to disambiguate LA/HA based on singular/plural agreement on the embedded verb, resulting in a $2\times2$ design (see Table \ref{tab:ENG2Stimuli}).\@

\begin{table}[htbp]
    \centering
    \begin{tabular}{|c|c|c|}
    \hline
        Sentence & Noun Type & Attachment \\\hline
        a & RC-only & HA \\\hline
        b &  RC-only & LA \\\hline
        c & SC & HA \\\hline
        d & SC & LA \\\hline
    \end{tabular}
    \caption{Summary of $2\times2$ design in the English Experiment 2.}
    \label{tab:ENG2Stimuli}
\end{table}

\begin{exe}
    \ex English Exp. 2 Stimuli \citep{grillo2015highs}
		\begin{xlist}
            \ex The picture of the son of the doctors that was having dinner is old.
            \ex The picture of the son of the doctors that were having dinner is old.
            \ex The car of the son of the doctors that was having dinner is old.
            \ex The car of the son of the doctors that were having dinner is old.
		\end{xlist}
        \label{exe:EnglishNounStimuli}
\end{exe}

Results from linear-mixed effect models for each LLM are again mixed, but generally in line with those in the first English experiment (Figure \ref{fig:E2}, Appendix \ref{sec:appa}).\@
For the bert-base model, we find a strong effect of Attachment Type, no effect of Noun Type, and no interaction.\@
These are compatible with bert strongly preferring LA items independently of the noun manipulation.\@
For the xlm model, we found a significant clause type effect, but no effect of attachment, nor an interaction.\
Finally, we again found a strong Attachment Type effect for the Roberta model, this time with no interaction with Noun Type.\@
This is the result of a strong preference for HA items across Noun Type conditions (Figure \ref{fig:E2}).\@

\section{Discussion and Further Work\footnote{\textnormal{ Anonymized scripts and data for all the experiments in this paper can be found at \href{https://osf.io/f85zh/?view\_only=3448344fd0244341a015a68246a25e47}{https://shorturl.at/n22lv.}}}}
\label{sec:FurtherWork}

In this work, we measured the difference in surprisal of locally ambiguous sentences at the point of disambiguation (the embedded verb) to determine whether a number of (monolingual and multilingual) LLMs learn human-like RC attachment preferences in Italian and English.\
Furthermore, we tested whether these preferences can be modulated by lexical factors in the matrix clause (Verb Type or Noun Type), which have been argued to be related to subtle differences between RCs and other constructions.\ 

For Italian, our results indicate that none of the models we tested exhibits any attachment preference at all, whether in line with the human results or not.\
However, we do observe high item-level variability, which should be an important focus for future studies.\
Even though we control for item-level lexical effects in our statistical models, because of this stark item-based variability we do note interesting (non statistically significant) tendencies in some of the models that beg for deeper inquiry in future work (see Appendix \ref{sec:appb}).\@
For instance, modulo some high surprisal LA items, the GePpetto model shows a general qualitative preference towards LA, in particular with perceptual verbs.\@

Notably, our statistical results are also somewhat in contrast with what previous work found for Spanish and Arabic \citep{davis2020recurrent, ISSA2024425}.\ 
However, \citet{davis2020recurrent} tested models with an LSTM architecture, while \citet{ISSA2024425} used prompting methods as opposed to the surprisal measurements used here.\@
Future work should then probe differences between architectures and tasks/measures more in depth.\@

English results across two experiments where more mixed.\
While some models did showcase some type of attachment preference, and at times verb and noun type effects on these preferences, these were not exactly in line with human data.\
For instance, while the bert-base model does show a slight preference for LA items, the roberta model shows a strong bias towards HA items, in contrast with the reported LA preference for English.
The mirrored behavior of bert and roberta across the two English experiments is also of note, and opens question for future comparisons --- as does the fact that surprisal values across models were slightly higher in the RC-only condition.\ 

Finally, beyond extending our investigation of RC attachment and Pseudorelatives to other languages \citep[e.g., Spanish;][]{aguilarSpanishNoDifferent2021}, richer insight into LLMs' linguistic knowledge will come from probing their ability to handle other factors known to affect RC disambiguation strategies in humans \cite[e.g., length;][]{hemforth2015relative}.

Overall, these results suggest a primary role for RC disambiguation in the study of LLMs' capabilities cross-linguistically, and strengthen the argument in favor of psycholinguistically motivated benchmarks for the rigorous evaluation of LLMs' abilities across languages.\@

\section*{Limitations}
In this paper we relied on experimental items available from two psycholinguistic studies of interest.
However, this meant that the number of items used in the paper is relatively low compared to the size of test sets in the LLM literature.\
Relatedly, the item-level variability observed in our results deserves further investigation.
Additionally, a limitation of comparing Italian to English is that in Italian surprisal is measured at the disambiguating verb, which varies across sets, but in English the disambiguating continuation is always measured on the \emph{was/were} contrast.\
Finally, in terms of comparison with previous literature, previous work found attachment preferences in English and Spanish with LSTMs, and in Arabic with a different subset of Transformer models.\
A better understanding of the relation between this past work and our results will come from testing similar constructions while keeping architectural (and task) details constant.\
Our work also limited its evaluation to Italian and English.\
Future work on RC attachment and noun/verb type effects should be extended to multiple languages with and without pseudo-relative constructions.

\bibliography{anthology,custom}
\bibliographystyle{aclNatbib}

\appendix
 \clearpage
 
\onecolumn
\section{Summary of LME Models}
\label{sec:appa}

\begin{table*}[htbp]
\begin{subtable}{1\textwidth}
\centering
\resizebox{1\textwidth}{!}{
\begin{tabular}{|l|r|r|r|r|r|}
\hline
                            & Estimate & Std. Error & df       & t-value & Pr(\textgreater{}|t|) \\ \hline
(Intercept)                 & 27.94089 & 0.82105    & 31.02296 & 34.031  & \textless{}2e-16      \\ \hline
Verb Type                   & -0.63548 & 0.50688    & 69.00000 & -1.254  & 0.214                 \\ \hline
Attachment Type             & -0.19745 & 0.50688    & 69.00000 & -0.390  & 0.698                 \\ \hline
Verb Type : Attachment Type & -1.34373 & 4.39996    & 69.00000 & -0.305  & 0.761                 \\ \hline
\end{tabular}
}
\label{tbl:alberto}
\caption{Alberto}
\end{subtable}
\begin{subtable}{1\textwidth}
\centering
\resizebox{1\textwidth}{!}{
\begin{tabular}{|l|r|r|r|r|r|}
\hline
                            & Estimate & Std. Error & df      & t-value & Pr(\textgreater{}|t|) \\ \hline
(Intercept)                 & 2.5262   & 0.6478     & 80.3745 & 3.900   & 0.000199              \\ \hline
Verb Type                   & -0.7534  & 0.8093     & 69.0000 & -0.931  & 0.355153              \\ \hline
Attachment Type             & 0.1703   & 0.8093     & 69.0000 & 0.210   & 0.833988              \\ \hline
Verb Type : Attachment Type & 1.2688   & 1.1446     & 69.0000 & 1.109   & 0.271492              \\ \hline
\end{tabular}
}
\label{tbl:geppetto}
\caption{GePpeTto}
\end{subtable}
\begin{subtable}{1\textwidth}
\centering
\resizebox{1\textwidth}{!}{
\begin{tabular}{|l|r|r|r|r|r|}
\hline
                            & Estimate & Std. Error & df      & t-value & Pr(\textgreater{}|t|) \\ \hline
(Intercept)                 & 28.1589  & 1.6636     & 32.7576 & 16.927  & \textless{}2e-16      \\ \hline
Verb Type                   & -0.4183  & 1.1124     & 69.0000 & -0.376  & 0.708                 \\ \hline
Attachment Type             & -0.5246  & 1.1124     & 69.0000 & -0.472  & 0.639                 \\ \hline
Verb Type : Attachment Type & 0.6105   & 1.5732     & 69.0000 & 0.388   & 0.699                 \\ \hline
\end{tabular}
}
\label{tbl:bert}
\caption{bert\_base\_multilingual\_case}
\end{subtable}
\begin{subtable}{1\textwidth}
\centering
\resizebox{1\textwidth}{!}{
\begin{tabular}{|l|r|r|r|r|r|}
\hline
                            & Estimate & Std. Error & df      & t-value & Pr(\textgreater{}|t|) \\ \hline
(Intercept)                 & 24.6305  & 2.0505     & 54.8460 & 12.012  & \textless{}2e-16      \\ \hline
Verb Type                   & 1.7987   & 2.2630     & 60.0000 & 0.795   & 0.430                 \\ \hline
Attachment Type             & 0.3184   & 2.2630     & 60.0000 & 0.141   & 0.889                 \\ \hline
Verb Type : Attachment Type & -0.4746  & 3.2003     & 60.0000 & -0.148  & 0.883                 \\ \hline
\end{tabular}
}
\label{tbl:xlm}
\caption{xlm-mlm-17-1280}
\end{subtable}
\begin{subtable}{1\textwidth}
\centering
\resizebox{1\textwidth}{!}{
\begin{tabular}{|l|r|r|r|r|r|}
\hline
                            & Estimate & Std. Error & df       & t-value & Pr(\textgreater{}|t|) \\ \hline
(Intercept)                 & 46.35441 & 3.29053    & 47.98074 & 14.087  & \textless{}2e-16      \\ \hline
Verb Type                   & 3.62229  & 3.11124    & 69.00000 & 1.164   & 0.248                 \\ \hline
Attachment Type             & 0.08279  & 3.11124    & 69.00000 & 0.027   & 0.979                 \\ \hline
Verb Type : Attachment Type & -1.34373 & 4.39996    & 69.00000 & -0.305  & 0.761                 \\ \hline
\end{tabular}
}
\label{tbl:rob}
\caption{xlm-roberta-large}
\end{subtable}
\caption{LMER Summary for all models in the Italian Experiment. Signif. codes:  0 `***' 0.001 ‘**’ 0.01 ‘*’ 0.05.}
 \label{tab:LinearMixedEffectITA}
\end{table*}

\begin{table*}[htbp]
\begin{subtable}{1\textwidth}
\centering
\begin{tabular}{|l|r|r|r|r|r|}
\hline
                            & Estimate & Std. Error & df      & t-value & Pr(\textgreater{}|t|) \\ \hline
(Intercept)                 & 19.9798  & 0.5074     & 55.6898 & 39.377  & \textless{}2e-16      \\ \hline
Verb Type                   & -1.7528  & 0.5243     & 69.0000 & -3.343  & 0.00134**             \\ \hline
Attachment Type             & -0.7160  & 0.5243     & 69.0000 & -1.366  & 0.17648               \\ \hline
Verb Type : Attachment Type & -0.5337  & 0.7414     & 69.0000 & -0.720  & 0.47407               \\ \hline
\end{tabular}
\label{tbl:bert}
\caption{bert\_base\_multilingual\_case}
\end{subtable}
\begin{subtable}{1\textwidth}
\centering
\begin{tabular}{|l|r|r|r|r|r|}
\hline
                            & Estimate & Std. Error & df      & t-value & Pr(\textgreater{}|t|) \\ \hline
(Intercept)                 & 27.7412  & 0.7469     & 45.2554 & 37.144  & \textless{}2e-16      \\ \hline
Verb Type                   & -0.9368  & 0.6790     & 69.0000 & -1.380  & 0.1721                \\ \hline
Attachment Type             & -1.3458  & 0.6790     & 69.0000 & -1.982  & 0.0515                \\ \hline
Verb Type : Attachment Type & 1.8169   & 0.9602     & 69.0000 & 1.892   & 0.0627                \\ \hline
\end{tabular}
\label{tbl:xlm}
\caption{xlm-mlm-17-1280}
\end{subtable}
\begin{subtable}{1\textwidth}
\centering
\begin{tabular}{|l|r|r|r|r|r|}
\hline
                            & Estimate & Std. Error & df      & t-value & Pr(\textgreater{}|t|) \\ \hline
(Intercept)                 & 25.2791  & 0.5831     & 53.7553 & 43.350  & \textless{}2e-16      \\ \hline
Verb Type                   & -1.7132  & 0.5907     & 69.0000 & -2.900  & 0.00499**             \\ \hline
Attachment Type             & 2.8251   & 0.5907     & 69.0000 & 4.783   & 9.46e-06***           \\ \hline
Verb Type : Attachment Type & 0.2826   & 0.8353     & 69.0000 & 0.338   & 0.73616               \\ \hline
\end{tabular}
\label{tbl:rob}
\caption{xlm-roberta-large}
\end{subtable}
\caption{LMER Summary for all models in the English Experiment 1. Signif. codes:  0 `***' 0.001 ‘**’ 0.01 ‘*’ 0.05.}
\label{tab:LinearMixedEffectEN1}
\end{table*}

\begin{table*}[htbp]
\begin{subtable}{1\textwidth}
\centering
\begin{tabular}{|l|r|r|r|r|r|}
\hline
                            & Estimate & Std. Error & df      & t-value & Pr(\textgreater{}|t|) \\ \hline
(Intercept)                 & 21.2704  & 0.4423     & 47.9400 & 48.094  & \textless 2e-16       \\ \hline
Noun Type                   & -0.5328  & 0.4179     & 69.0000 & -1.275  & 0.207                 \\ \hline
Attachment Type             & -2.4803  & 0.4179     & 69.0000 & -5.935  & 1.06e-07***           \\ \hline
Noun Type : Attachment Type & -0.5808  & 0.5911     & 69.0000 & -0.983  & 0.329                 \\ \hline
\end{tabular}
\label{tbl:bert}
\caption{bert\_base\_multilingual\_case}
\end{subtable}
\begin{subtable}{1\textwidth}
\centering
\begin{tabular}{|l|r|r|r|r|r|}
\hline
                            & Estimate & Std. Error & df      & t-value & Pr(\textgreater{}|t|) \\ \hline
(Intercept)                 & 24.6088  & 0.9001     & 51.2856 & 27.339  & \textless 2e-16       \\ \hline
Noun Type                   & 1.9332   & 0.8871     & 69.0000 & 2.179  & 0.0327*               \\ \hline
Attachment Type             & 1.5127   & 0.8871     & 69.0000 & 1.705   & 0.0926                \\ \hline
Noun Type : Attachment Type & -1.1707  & 1.2545     & 69.0000 & -0.933  & 0.3540                \\ \hline
\end{tabular}
\label{tbl:xlm}
\caption{xlm-mlm-17-1280}
\end{subtable}
\begin{subtable}{1\textwidth}
\centering
\begin{tabular}{|l|r|r|r|r|r|}
\hline
                            & Estimate & Std. Error & df      & t-value  & Pr(\textgreater{}|t|) \\ \hline
(Intercept)                 & 27.1706  & 0.5632     & 44.8643 & 48.243   & \textless 2e-16       \\ \hline
Noun Type                   & -0.5667  & 0.5089     & 69.0000 & -1.114   & 0.269274              \\ \hline
Attachment Type             & 2.0496   & 0.5089     & 69.0000 & 4.028    & 0.000143***           \\ \hline
Noun Type : Attachment Type & 0.3905   & 0.7197     & 69.0000 & 0.589182 & 0.589182              \\ \hline
\end{tabular}
\label{tbl:rob}
\caption{xlm-roberta-large}
\end{subtable}
\caption{LMER Summary for all models in the English Experiment 2. Signif. codes:  0 `***' 0.001 ‘**’ 0.01 ‘*’ 0.05}
\label{tab:LinearMixedEffectEN2}
\end{table*}

\clearpage
\twocolumn

\section{Preferences as Categorical Pairwise Comparisons}
\label{sec:appb}

\begin{figure*}[htbp]
    \includegraphics[width=1.0\linewidth]{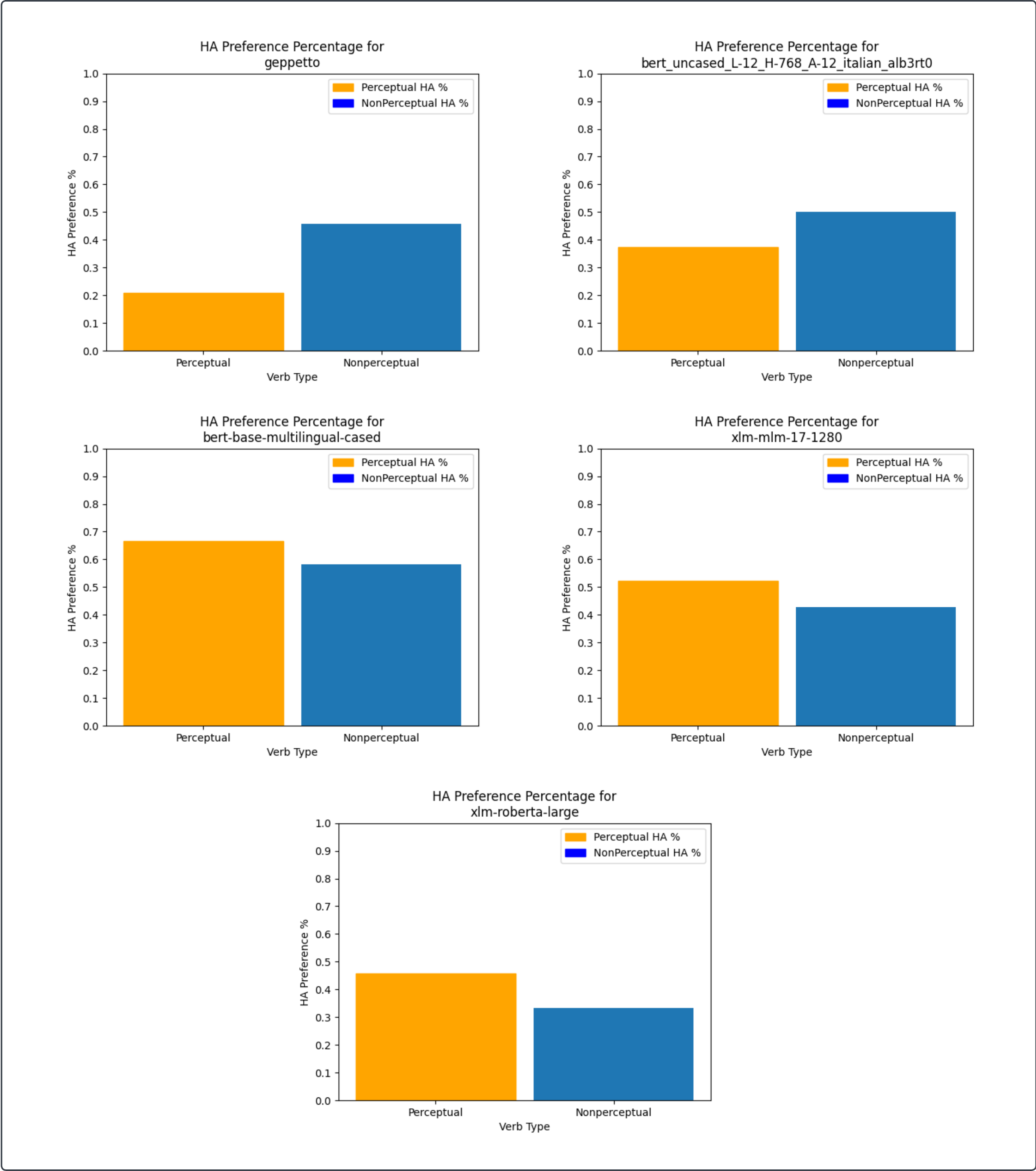}
    \caption{Proportion of HA vs. LA in the Italian Experiment, derived from categorical pairwise comparisons within sets.}
         \label{fig:ITAHAFULL}
\end{figure*}

\begin{figure*}[htbp]
     \begin{subfigure}[t]{0.5\textwidth}
 \centering
     \includegraphics[width=1.0\linewidth]{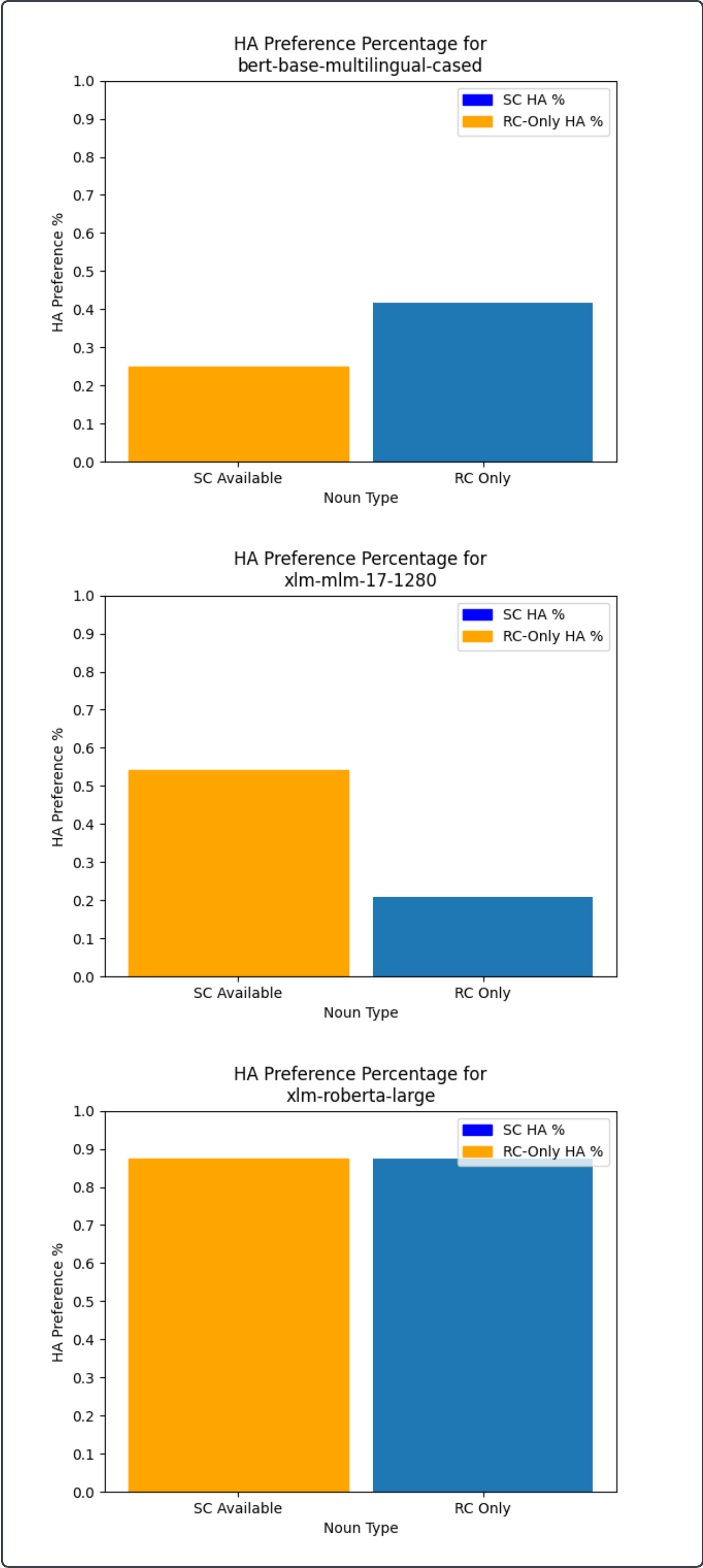}
     \caption{}
     \label{fig:ENHA1}
        \end{subfigure}%
    ~ 
         \begin{subfigure}[t]{0.5\textwidth}
 \centering
     \includegraphics[width=1.0\linewidth]{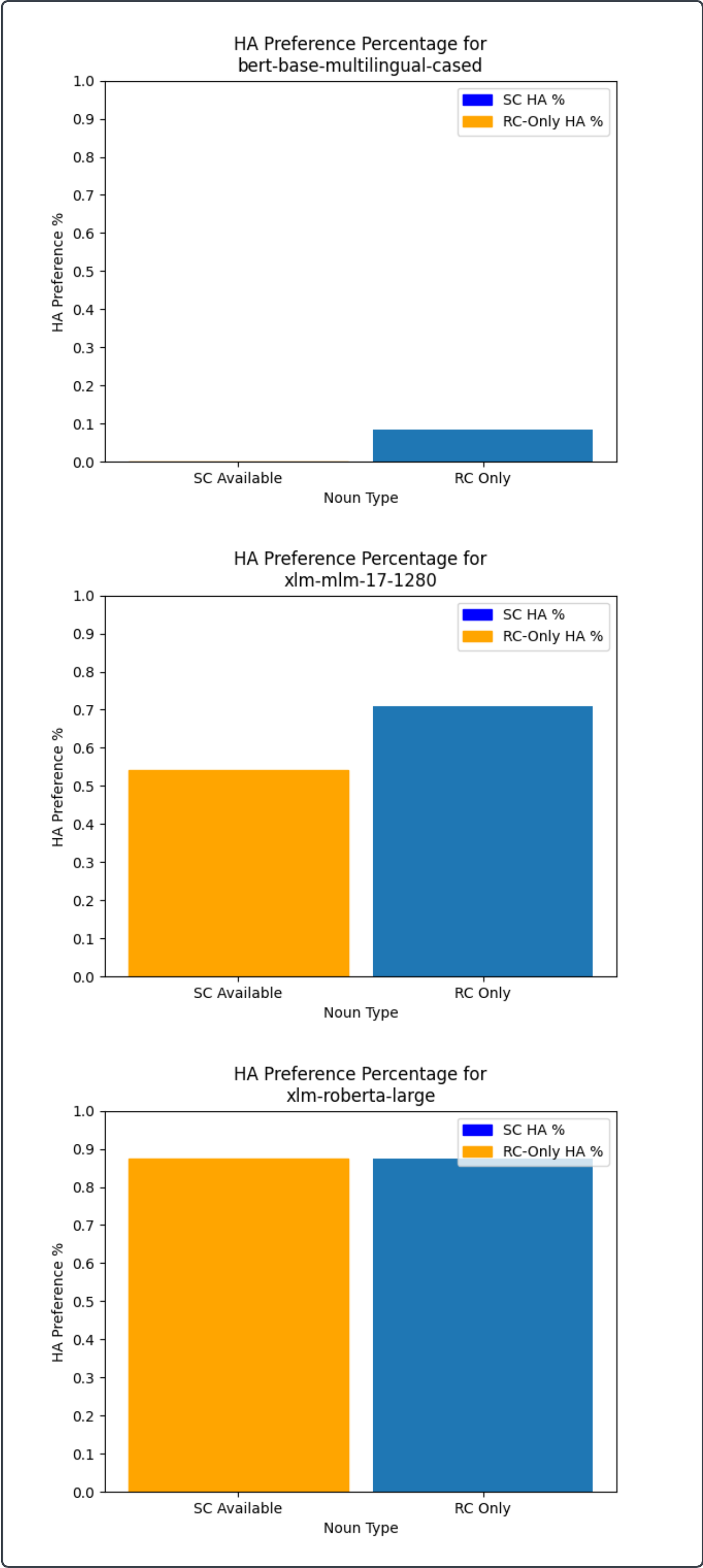}
        \caption{}
     \label{fig:EHA2}
       \end{subfigure}%
                 \caption{Proportion of HA vs. LA in the English Experiment 1 (a) and 2 (b), derived from categorical pairwise comparisons within sets.}
                              \label{fig:EnglishHAFULL}
   \end{figure*}

Following standard practices in psycholinguistics, for the paper's core analyses statistical robustness of the effects/contrasts has been determined by running linear-mixed effects models using the original distribution of surprisal values over items.\
This is also consistent with what is done with human data during (for instance) online tasks involving locally ambiguous sentences like the ones we used.\
However, a qualitative understanding of model's trend, in line with results from human participants from forced choices tasks targeting globally ambiguous sentences, can be achieved by coding a model's preference for HA/LA categorically for each item pair in a set \citep{davis2020recurrent}.\
That is, in each set items can be paired by keeping Verb Type/Noun Type consistent.
Then, if surprisal for the LA disambiguated item was lower than the surprisal for the HA disambiguated item, attachment is coded as LA.\
See Example \ref{exe:AttachmentITA} for a summary of this coding approach across the experiments in this paper, and Figure \ref{fig:ITAHAFULL} and Figure \ref{fig:EnglishHAFULL} for a visualization of model preferences given this kind of coding schema.
Note that the statistical significance of these contrasts is still as discussed previously in the paper and summarized in Tables \ref{tab:LinearMixedEffectITA}, \ref{tab:LinearMixedEffectEN1}, and \ref{tab:LinearMixedEffectEN2}.

\begin{exe}
    \ex Interpretation of Pairwise comparisons for each experiment
		\begin{xlist}
    \item \texttt{Attachment Preference} $\leftarrow$ \texttt{LOW} if Verb Surprisal(a) $>$ Verb Surprisal(b)
    \item \texttt{Attachment Preference} $\leftarrow$ \texttt{HIGH} if Verb Surprisal(a) $<$ Verb Surprisal(b)
    \item \texttt{Attachment Preference} $\leftarrow$ \texttt{LOW} if Verb Surprisal(c) $>$ Verb Surprisal(d)
    \item \texttt{Attachment Preference} $\leftarrow$ \texttt{HIGH} if Verb Surprisal(c) $<$ Verb Surprisal(d)
    \end{xlist}
    \label{exe:AttachmentITA}
\end{exe}

\end{document}